\documentclass[12pt]{article}

\usepackage{sbc-template}
\usepackage{graphicx,url}
\usepackage[utf8]{inputenc}
\usepackage[brazil]{babel}

\usepackage{amsmath} 
\usepackage{txfonts} 

\usepackage[linesnumbered,vlined,ruled,boxed,commentsnumbered]{algorithm2e}
  \usepackage{algpseudocode}
  \usepackage{bm} 

\usepackage{threeparttable}
\usepackage{multirow}
\usepackage{bbm}

\usepackage{color, colortbl}

\usepackage{verbatim}

\newcommand{\tref}[1]{Table \ref{#1}}

\newcommand{\eref}[1]{Eq. \ref{#1}}

\usepackage{cleveref}

\usepackage{multirow,booktabs,color,soul,threeparttable}
\definecolor{hl}{rgb}{0.75,0.75,0.75}
\sethlcolor{hl}

\usepackage{graphicx}

\usepackage{scalefnt}

\usepackage{float}
\usepackage{subfigure} 

\title{A Non-Dominated Sorting Evolutionary Algorithm Updating When Required}

\author{Lucas R. C. Farias\inst{1,2,3}, Abimael J. F. Santos\inst{1}, Matheus R. B. Nobre\inst{1}}
  
\address{
  Centro de Informática\\
  Universidade Federal de Pernambuco (UFPE) -- Recife, PE -- Brazil
  \email{\{lrcf, ajfs, mrbn\}@cin.ufpe.br}
  \nextinstitute
  CESAR School - Recife, Pernambuco -- Brazil
  \email{lrcf@cesar.school}
  \nextinstitute
  Departamento de Ciência da Computação\\
  Universidade Católica de Pernambuco (UNICAP) -- Recife, PE -- Brazil
  \email{lucas.farias@unicap.br}
}

\begin{document} 

\maketitle

\begin{abstract}
The NSGA-III algorithm relies on uniformly distributed reference points to promote diversity in many-objective optimization problems. However, this strategy may underperform when facing irregular Pareto fronts, where certain vectors remain unassociated with any optimal solutions. While adaptive schemes such as A-NSGA-III address this issue by dynamically modifying reference points, they may introduce unnecessary complexity in regular scenarios. This paper proposes NSGA-III with Update when Required (NSGA-III-UR), a hybrid algorithm that selectively activates reference vector adaptation based on the estimated regularity of the Pareto front. Experimental results on benchmark suites (DTLZ1–7, IDTLZ1–2), and real-world problems demonstrate that NSGA-III-UR consistently outperforms NSGA-III and A-NSGA-III across diverse problem landscapes.
\end{abstract}

\section{Introduction}

Multi-objective optimization problems (MOPs) are defined by the simultaneous optimization of two or more conflicting objective functions, where the goal is to approximate the Pareto optimal front (POF) that captures the trade-offs among the objectives. In this context, evolutionary algorithms have gained prominence due to their inherent ability to explore diverse regions of the solution space and maintain a population of non-dominated solutions across generations \cite{deb2014evolutionary, ishibuchi2017performance}.

As the number of objectives increases, leading to many-objective optimization problems (MaOPs), conventional dominance-based algorithms face significant challenges related to convergence and diversity preservation. NSGA-III has emerged as a leading approach for MaOPs, using a set of uniformly distributed reference points in the objective space to guide selection pressure and ensure solution dispersion \cite{deb2014evolutionary}. However, its reliance on fixed reference vectors poses a limitation when dealing with problems characterized by irregular or non-convex Pareto fronts, where some reference points may never be associated with any solution, while others may attract multiple candidates \cite{jain2013evolutionary}.

Several adaptive extensions have been proposed to address this, including A-NSGA-III \cite{jain2013evolutionary}, which dynamically adjusts the reference points through inclusion and exclusion operations based on the population's distribution. Although these approaches introduce flexibility, they may inadvertently distort the uniformity of reference vectors, particularly when adaptation is applied in contexts where the original configuration remains adequate, such as problems with regular POFs.

Recent studies have highlighted the importance of context-awareness in adaptive optimization, where the decision to alter the reference vectors should consider the intrinsic shape and regularity of the POF \cite{de2022decomposition}. In particular, Farias and Araújo proposed a method to assess the regularity of the objective space via the Spreading Index (SI), enabling the dynamic selection of hyperparameter configurations. Their findings reinforce that adaptation should be context-driven and applied only when beneficial, an insight aligned with the No Free Lunch Theorem.

Building upon these foundations, this paper introduces a novel many-objective algorithm, termed NSGA-III with Update when Required (NSGA-III-UR). The core idea is to integrate a decision mechanism based on the SI metric and a threshold model to determine whether the reference vectors should be updated during the evolutionary process. This mechanism enables the algorithm to switch between standard NSGA-III behavior and an adapted A-NSGA-III scheme, depending on the estimated regularity of the POF.

The main contributions of this work are summarized as follows:
\begin{enumerate}
 \item We propose a context-aware strategy for updating reference vectors in NSGA-III, guided by the spreading index and applied only when the POF is irregular.
 \item We present a hybrid framework that preserves the simplex-lattice structure for regular problems while enabling adaptive relocation of reference vectors for irregular ones.
 \item We conduct extensive experiments on benchmark problems (DTLZ1–7, IDTLZ1–2) and real-world scenarios (multi-objective knapsack and water resource planning), demonstrating that NSGA-III-UR consistently outperforms NSGA-III and A-NSGA-III across diverse POF shapes.
 \end{enumerate}

The remainder of this paper is organized as follows. Section 2 presents the related works. Section 3 describes the proposed NSGA-III-UR in detail. Section 4 shows the validation on benchmarks and real-world problems. Section 5 draws some conclusions and presents several suggestions for future lines of research.

\section{Related Works}

Many-objective optimization problems (MaOPs) pose significant challenges for conventional multi-objective evolutionary algorithms (MOEAs), particularly due to the curse of dimensionality, loss of selection pressure, and difficulties in maintaining diversity across irregular Pareto fronts (POFs) \cite{deb2014evolutionary, ishibuchi2017performance}. A broad spectrum of MOEA frameworks has been proposed to address these issues, including Pareto-dominance-based, decomposition-based, and indicator-based strategies.

Among the most prominent Pareto-based algorithms, NSGA-II and NSGA-III have gained widespread adoption. NSGA-II employs a crowding distance mechanism to preserve diversity, yielding satisfactory results in problems with up to three objectives. However, as the number of objectives increases, its effectiveness diminishes due to the reduced dominance discrimination among individuals \cite{awad2022portfolio}. NSGA-III mitigates this by introducing a reference point-based selection mechanism, enabling better distribution in higher-dimensional objective spaces. Nevertheless, its fixed reference vector design can lead to inefficiencies in problems with irregular or disconnected POFs, where certain regions of the objective space remain underrepresented \cite{jain2013evolutionary, ishibuchi2016performance}.

To overcome these limitations, several adaptive extensions have been developed. A-NSGA-III \cite{jain2013evolutionary} introduces reference point adaptation via inclusion and exclusion operations, allowing the algorithm to reshape its vector distribution dynamically. While effective, this adaptation may compromise the uniformity of the simplex-lattice structure in problems with regular POFs, resulting in unnecessary complexity and possible performance degradation.

Complementary strategies have emerged in the decomposition-based paradigm. MOEA/D and its derivatives, such as MOEA/D-LNA and MOEA/D-AWS \cite{nuh2021performance, junqueira2022moead}, decompose the problem into scalar subproblems and employ neighborhood-based or external archive-based mechanisms to adapt weight vectors. Recent work on MaOEA/D-AEW \cite{sun2024maoead} integrates adaptive external populations to enhance exploitation capabilities while preserving exploratory potential. These methods perform strongly on regular and irregular fronts but are often sensitive to weight vector configurations.

Another direction of advancement lies in improving performance assessment. Traditional indicators like Inverted Generational Distance (IGD) and Hypervolume (HV), although widely adopted, present limitations in reflecting the diversity and convergence of solutions in high-dimensional or non-convex landscapes \cite{nuh2021performance}. This has motivated the development of problem-specific performance metrics and threshold-based criteria to characterize the search dynamics better and adjust algorithm behavior accordingly.

Context-aware adaptation has recently gained traction as a mechanism for dynamic decision-making during evolution. Farias and Araújo \cite{de2022decomposition} proposed a framework in which the regularity of the POF is estimated using the Spreading Index (SI), enabling conditional activation of adaptation procedures. This method supports the premise that not all problems benefit equally from adaptive mechanisms; in regular problems, static reference vectors may suffice, while irregular problems require more flexible structures.

Despite these advances, a unified approach that selectively enables adaptation only when necessary—thus avoiding redundant computation and preserving structural integrity—is still underexplored. This paper builds upon the aforementioned insights by proposing NSGA-III-UR, a hybrid algorithm incorporating context-aware adaptation through regularity detection. In doing so, it seeks to reconcile the strengths of fixed and adaptive reference point schemes, delivering competitive performance across a wide range of POF geometries and problem instances.

\section{NSGA-III with Update when Required}

The NSGA-III with Update when Required (NSGA-III-UR) is a hybrid many-objective evolutionary algorithm that conditionally adapts its reference vectors based on the geometric regularity of the Pareto-optimal front (POF). The algorithm integrates the selection framework of NSGA-III with the adaptive reference vector scheme of A-NSGA-III, augmented by a statistical metric that triggers adaptation only when necessary.

\begin{figure}[H]
        \centering
        \includegraphics[width=0.56\linewidth]{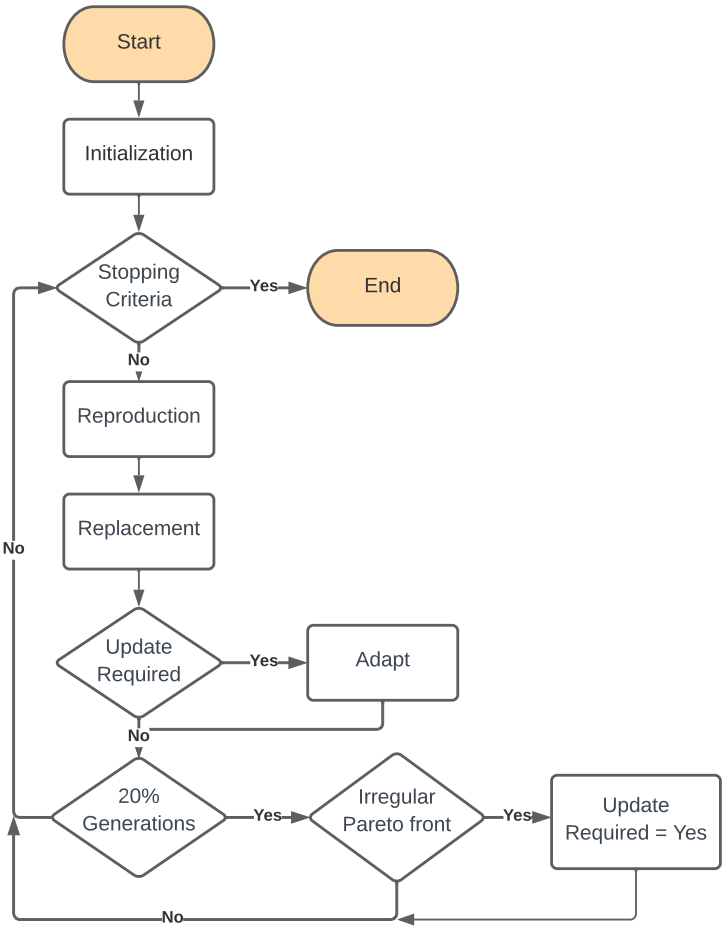}
        \caption{Flowchart of execution of the NSGA-III-UR.}
        \label{fig:flux-NSGAIIIUR}
\end{figure}

Figure~\ref{fig:flux-NSGAIIIUR} illustrates the high-level execution flow of NSGA-III-UR, comprising six core components: (i) initialization, (ii) reproduction via genetic operators, (iii) environmental selection using reference vectors and dominance sorting, (iv) optional adaptation of reference vectors, (v) condition-based activation of adaptation, and (vi) termination by a predefined criterion. We now describe the core elements of the proposed approach.

\subsection{NSGA-III Selection Framework}

NSGA-III is a Pareto-based many-objective evolutionary algorithm that maintains population diversity by projecting individuals onto a set of uniformly distributed reference vectors in the objective space \cite{deb2014evolutionary}. These vectors are generated via the Das and Dennis method \cite{das1998normal}, which discretizes a unit simplex into a lattice of $N = \binom{m + H - 1}{H}$ points, where $m$ is the number of objectives and $H$ the granularity level.

To perform environmental selection, the algorithm (i) normalizes population members using the ideal point derived from the current generation, (ii) associates each individual with the nearest reference vector, and (iii) applies a niche preservation operation that selects the solution closest to each reference vector. This mechanism promotes diversity but may degrade when the POF deviates significantly from the idealized simplex shape.

\subsection{A-NSGA-III Adaptation Scheme}

A-NSGA-III introduces an adaptive mechanism for reference vector relocation, guided by the distribution of individuals across the current generation \cite{jain2013evolutionary}. The algorithm performs two operations: (i) inclusion, which adds new reference vectors in the vicinity of overpopulated vectors using a local simplex design, and (ii) exclusion, which removes reference vectors unassociated with any solution.

While effective on irregular POFs, this strategy can introduce redundancy or loss of structure on regular POFs. Based on an online diagnostic of POF regularity, NSGA-III-UR mitigates this by invoking adaptation only when required.

\subsection{Update When Required: Context-Aware Adaptation Trigger}

To dynamically determine whether reference vector adaptation is beneficial, we incorporate the Spreading Index (SI) and a threshold function originally proposed by Farias and Araújo \cite{de2022decomposition}. The SI quantifies the geometric irregularity of the fitness distribution within the population as follows:

\begin{equation}
\label{eq:regularityMetric}
SI(\mathbf{f'}, h) = \frac{\sqrt{\sum_{i=1}^{N} \sum_{j=1}^{m} f_{ij}^2}}{h},
\end{equation}

where $f_{ij}$ denotes the $j$-th normalized objective of the $i$-th individual, and $h$ is a normalization factor (empirically set to 4).

A cubic regression threshold function defines a critical value for $SI$ given the number of objectives $m$:

\begin{algorithm}
\caption{NSGA-III-UR}
\label{alg:framework}
    
    $start \gets 0.2$; $mode \gets 0$;
    
    Calculate the number of reference points, $N$, to place on the hyperplane;
    
    Generate the initial population $\boldsymbol{P}$ at random;
    
    \While{$Gen < Gen_{max}$}
        {
        
        Select two set of parents, $\boldsymbol{P_1}$ and $\boldsymbol{P_2}$, using the tournament method;
        
        Apply the recombination and mutation operators between $\boldsymbol{P_1}$ and $\boldsymbol{P_2}$;

        Realize the non-dominated population sorting;
        
        Normalize the population members;
        
        Associate the population member with the reference points;
        
        Apply the niche preservation;
        
        Keep the niche obtained solutions for the next generation;
        
        \If {$mode$ == 1}
            {
                Perform the reference vectors adaptive like A-NSGA-III
            }
        \If{$Gen/Gen_{max} == start$}
            {
                Calculate the $spreading$ $index$ using \eref{eq:regularityMetric};
        
                Calculate the $threshold$ using \eref{eq:limiar};
                
                \If{$spreading$ $index > threshold$}
                {
                    $mode \gets 1$;
                }
            }
        }
    \textbf{return} $\boldsymbol{P}$;
\end{algorithm}

\begin{equation}
\label{eq:limiar}
\text{threshold}(m) = -0.00001989 m^3 + 0.0002034 m^2 + 0.03376 m - 0.2373.
\end{equation}

If $SI > \text{threshold}(m)$, the POF is deemed irregular and adaptation is enabled. Otherwise, NSGA-III proceeds without modifying its reference vectors. This decision is evaluated once, at a predefined stage of the evolutionary process (e.g., at 20\% of the total evaluations), to avoid frequent mode switching and preserve computational efficiency.

\begin{figure}[htbp]
\centering
\includegraphics[width=0.6\linewidth]{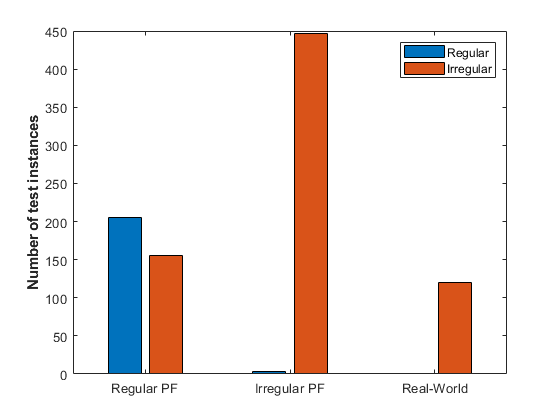}
\caption{Classification accuracy of POF regularity by NSGA-III-UR across benchmark and real-world problems.}
\label{fig:regularidadePorCategoria}
\end{figure}

Figure~\ref{fig:regularidadePorCategoria} summarizes the classification success of this mechanism across multiple benchmark and real-world scenarios.

\subsection{Algorithm Framework}

Algorithm~\ref{alg:framework} presents the pseudocode of NSGA-III-UR. The algorithm begins by initializing the reference vectors and generating the initial population, followed by selection, reproduction, and environmental replacement operations. The main novelty lies in lines 14–18: the POF regularity is assessed via $SI$ and used only to activate the adaptive reference vector mechanism (A-NSGA-III) when required.

\begin{itemize}
\item Lines 2–11: Standard NSGA-III execution: population initialization, reproduction, non-dominated sorting, normalization, reference association, and niche preservation.
\item Lines 12–13: Conditional execution of adaptive vector update in the A-NSGA-III, if flagged by the mode variable.
\item Lines 14–18: At the evaluation threshold, compute $SI$ and decide whether to activate adaptation mode.
\end{itemize}

\subsection{Computational Complexity}

The overall complexity per generation of NSGA-III-UR matches that of A-NSGA-III. This is dominated by either the sorting operation, with complexity $\mathcal{O}(N^2 \log_{M-2} N)$, or the reference vector adaptation, with complexity $\mathcal{O}(N^2 M)$, whichever is greater.

\section{Validation of the Algorithm}

This Section presents the experimental validation of the proposed NSGA-III-UR algorithm. We detail the benchmark problems and real-world instances used, describe the experimental setup and parameter configurations, define the performance metrics adopted, and analyze the obtained results through statistical comparisons.

\subsection{Experimental Setup}
\label{sec:experimentalSetup}

To evaluate the performance and generalizability of NSGA-III-UR, we conducted comparative experiments on a diverse set of benchmark and real-world multi-objective problems. The benchmark suite includes the well-known DTLZ1–7 \cite{deb2005scalable} and IDTLZ1–2 \cite{jain2013evolutionary} problem families. These problems vary in the geometric structure of the Pareto-optimal front (POF), including regular (linear, concave) and irregular (degenerate, disconnected, inverted) shapes. Additionally, we included two real-world optimization scenarios: the multi-objective knapsack problem (MOKP) \cite{zitzler1999multiobjective} and a water resource planning (WRP) problem \cite{de2022decomposition}.

\tref{problemSettings} summarizes the characteristics of each problem, including the number of objectives, decision variables, and POF type. All algorithms were run for a maximum of 60,000 function evaluations. Population sizes were set to 120 for problems with up to four objectives and 105 for five-objective instances.

The NSGA-III-UR was implemented in MATLAB using the PlatEMO framework \cite{tian2017platemo}, which also provided the baseline implementations of NSGA-III and A-NSGA-III. Experiments were executed on a machine equipped with an Intel Core i9-9900KF CPU (3.60GHz, eight cores) and 32 GB of RAM.

We used the parameter settings recommended by the literature for the compared algorithms. Variation operators included simulated binary crossover (SBX) and polynomial mutation \cite{deb2014evolutionary}, with distribution indices of 20. Crossover and mutation probabilities were fixed at 1.0 and $1/d$, respectively, where $d$ is the number of decision variables. For the MOKP, one-point crossover and bitwise mutation were used.

\begin{table}[ht]
\caption{Settings of the number of the objectives and decision variables for each test problem.}
\label{problemSettings}
\centering
\begin{tabular}{@{}ccccc@{}}
\hline
\begin{tabular}[c]{@{}c@{}}Benchmark\\ Problem\end{tabular} & \textit{\begin{tabular}[c]{@{}c@{}}Objectives \\ ($m$)\end{tabular}} & \textit{\begin{tabular}[c]{@{}c@{}}Variables\\ ($d$)\end{tabular}} & \begin{tabular}[c]{@{}c@{}}Pareto\\ front\end{tabular} \\ \hline
\multicolumn{4}{c}{Regular Pareto front} \\ \hline
DTLZ1 & 3-5 & $m$-1+5 & Linear \\
DTLZ2-4 & 3-5 & $m$-1+10 & Concave \\ \hline
\multicolumn{4}{c}{Irregular Pareto front} \\ \hline
DTLZ5-6 & 3-5 & $m$-1+10 & Degenerate \\
DTLZ7 & 3-5 & $m$-1+20 & Disconnected \\
IDTLZ1& 3-5 & $m$-1+5 & Inverted \\
IDTLZ2& 3-5 & $m$-1+10 & Inverted \\ \hline
\multicolumn{4}{c}{Real-World Problem} \\ \hline
MOKP & 3-5 & 250 & Real-world \\
WRP & 5 & 3 & Real-world \\ \hline
\end{tabular}
\end{table}

The metrics for assessing the performance of each algorithm are the inverted generational distance (IGD) \cite{bosman2003balance} and the Hypervolume (HV) \cite{zitzler1999multiobjective}, defined as:

\textbf{IGD}: Let $\boldsymbol{P^*}$ be a set of uniformly distributed reference points on the PF and $\boldsymbol{P}$ be the set of solutions. The IGD value of $\boldsymbol{P}$ can be defined below:
    \begin{equation}
    \label{inverseGenerationalDistance}
    IGD(\boldsymbol{P},\boldsymbol{P^*}) = \frac{\sum^{}_{\boldsymbol{v} \in \boldsymbol{P^*}} d(\boldsymbol{v},\boldsymbol{P})} {\vert \boldsymbol{P^*}\vert},
    \end{equation}
where $d$($\boldsymbol{v}$,$\boldsymbol{P}$) is the Euclidean distance from point $\boldsymbol{v}$ $\in$ $\boldsymbol{P^*}$ to its nearest point in $\boldsymbol{P}$. $\vert\boldsymbol{P^*}\vert$ is the cardinality of $\boldsymbol{P^*}$. The computational experiments use 10,000 reference points \cite{tian2017platemo}; only for the water resource planning problem (WRP) the reference set contains, 2429 solutions \cite{de2022decomposition}. The smaller the IGD the better the quality of $\boldsymbol{P}$ for approximating the whole PF.

\textbf{HV}: Given a reference point $\boldsymbol{z}^r$ = ($z_1^r$ ...  $z_n^r$)$^T$ dominated by all Pareto-optimal solutions, the HV of a set of solutions $\boldsymbol{P}$ is defined as the volume of the objective space dominated by all solutions in $\boldsymbol{P}$, bounded by  $\boldsymbol{z}^r$:
    \begin{equation}
    \label{eq:hypervolume}
    HV(\boldsymbol{P}, \boldsymbol{z}^r) = Vol(\bigcup_{\boldsymbol{p} \in P} [f^{\boldsymbol{p}}_1, z_1^r] \times ... \times [f^{\boldsymbol{p}}_m, z_m^r]),
    \end{equation}
where $Vol(\cdot)$ denotes the Lebesgue measure. The larger HV the better the approximation quality of $\boldsymbol{P}$. We use a reference point 10\% higher than the upper bound of the PF for our experiments. To reduce the computational complexity of determining HV if $m > 4$, we use the Monte Carlo method with 1,000,000 sampling points to approximate its value.

The tests were run 30 times independently, and the mean and standard deviation of each metric value were recorded. The Mann–Whitney U test with a significance level of 0.05 is adopted to perform statistical analysis on the experimental results, where the symbols "+", "-" and "$\approx$" indicate that the result by another MOEA is significantly better, significantly worse and statistically similar to that obtained by NSGA-III-UR, respectively.

\begin{table}[ht]
\scalefont{0.82}
\renewcommand{\arraystretch}{1.2}
\centering
\caption{IGD mean values obtained by NSGA-III, A-NSGA-III, and NSGA-III-UR on DTLZ1-7 and IDTLZ1-2 with 3 to 5 objectives.}
\begin{tabular}{@{}ccccc@{}}
\toprule
Problem&$M$&NSGA-III&A-NSGA-III&NSGA-III-UR\\
\midrule
\multirow{3}{*}{DTLZ1}&3&\hl{1.7691e-2 (1.08e-4) $\approx$}&2.1990e-2 (4.93e-3) $-$&1.7843e-2 (9.13e-4)\\
&4&4.1368e-2 (1.95e-4) $\approx$&4.5010e-2 (4.38e-3) $-$&\hl{4.1354e-2 (2.28e-4)}\\
&5&6.2975e-2 (1.38e-4) $\approx$&7.0294e-2 (6.76e-3) $-$&\hl{6.2975e-2 (1.16e-4)}\\
\hline
\multirow{3}{*}{DTLZ2}&3&\hl{4.6738e-2 (3.40e-6) $+$}&4.7935e-2 (1.06e-3) $-$&4.7290e-2 (7.57e-4)\\
&4&\hl{1.2119e-1 (1.91e-5) $\approx$}&1.2268e-1 (1.38e-3) $-$&1.2137e-1 (8.09e-4)\\
&5&1.9561e-1 (4.18e-5) $\approx$&2.2732e-1 (5.75e-2) $-$&\hl{1.9560e-1 (2.86e-5)}\\
\hline
\multirow{3}{*}{DTLZ3}&3&\hl{4.9245e-2 (1.43e-3) $\approx$}&5.8158e-2 (6.40e-3) $-$&4.9291e-2 (2.80e-3)\\
&4&\hl{1.2833e-1 (1.16e-2) $\approx$}&2.1792e-1 (2.53e-1) $-$&1.3667e-1 (1.12e-2)\\
&5&4.9820e-1 (5.26e-1) $\approx$&6.8560e-1 (8.78e-1) $-$&\hl{2.2999e-1 (6.80e-2)}\\
\hline
\multirow{3}{*}{DTLZ4}&3&1.1268e-1 (1.71e-1) $-$&1.2982e-1 (1.87e-1) $-$&\hl{9.6583e-2 (1.51e-1)}\\
&4&\hl{1.5423e-1 (1.01e-1) $\approx$}&1.8827e-1 (1.35e-1) $-$&1.5441e-1 (1.01e-1)\\
&5&2.5274e-1 (1.06e-1) $\approx$&\hl{2.1446e-1 (6.12e-2) $+$}&2.2737e-1 (8.24e-2)\\
\hline
\multirow{3}{*}{DTLZ5}&3&9.4496e-3 (1.20e-3) $-$&8.3865e-3 (8.66e-4) $\approx$&\hl{8.1736e-3 (6.73e-4)}\\
&4&4.3564e-2 (6.25e-3) $-$&4.1140e-2 (9.95e-3) $\approx$&\hl{3.8712e-2 (8.79e-3)}\\
&5&1.2867e-1 (4.57e-2) $\approx$&1.1656e-1 (3.83e-2) $\approx$&\hl{1.1384e-1 (3.78e-2)}\\
\hline
\multirow{3}{*}{DTLZ6}&3&1.4867e-2 (2.13e-3) $-$&\hl{1.0705e-2 (1.08e-3) $\approx$}&1.1317e-2 (1.61e-3)\\
&4&\hl{6.9152e-2 (2.72e-2) $\approx$}&7.1148e-2 (1.81e-2) $\approx$&7.4189e-2 (2.49e-2)\\
&5&2.4592e-1 (8.63e-2) $\approx$&2.5991e-1 (1.05e-1) $\approx$&\hl{2.3952e-1 (6.36e-2)}\\
\hline
\multirow{3}{*}{DTLZ7}&3&6.5054e-2 (2.10e-3) $-$&\hl{6.3170e-2 (1.86e-3) $\approx$}&6.3360e-2 (2.18e-3)\\
&4&1.8998e-1 (3.82e-2) $\approx$&1.8977e-1 (3.56e-2) $\approx$&\hl{1.8708e-1 (3.68e-2)}\\
&5&3.4837e-1 (2.34e-2) $\approx$&\hl{3.4807e-1 (2.12e-2) $\approx$}&3.5316e-1 (2.45e-2)\\
\hline
\multirow{3}{*}{IDTLZ1}&3&2.6154e-2 (6.70e-4) $-$&1.9372e-2 (2.75e-4) $\approx$&\hl{1.9301e-2 (1.23e-4)}\\
&4&7.6193e-2 (3.36e-3) $-$&5.7415e-2 (2.63e-3) $\approx$&\hl{5.6314e-2 (2.22e-3)}\\
&5&1.0284e-1 (5.51e-3) $-$&8.8557e-2 (2.72e-2) $\approx$&\hl{8.7231e-2 (1.27e-2)}\\
\hline
\multirow{3}{*}{IDTLZ2}&3&\hl{6.4105e-2 (2.71e-3) $\approx$}&6.5092e-2 (3.23e-3) $\approx$&6.5220e-2 (4.47e-3)\\
&4&1.5087e-1 (4.60e-3) $\approx$&1.5140e-1 (1.26e-2) $\approx$&\hl{1.4971e-1 (9.50e-3)}\\
&5&2.6673e-1 (1.76e-2) $-$&2.7701e-1 (2.48e-2) $-$&\hl{2.5699e-1 (1.90e-2)}\\
\hline
\multicolumn{2}{c}{$+/-/\approx$}&1/9/17&1/12/14&\\
\bottomrule
\end{tabular}
\label{IGD_results}
\end{table}

\subsection{Results and Discussion}

Table 2 presents the IGD results across the 27 benchmark instances. NSGA-III-UR performed best in 15 cases, showing superior generalization to both regular (DTLZ 1–4) and irregular (DTLZ 5–7, IDTLZ 1–2) POFs. Specifically, it outperformed NSGA-III in 9 cases and A-NSGA-III in 12, reinforcing the effectiveness of its context-aware adaptation strategy. In a few cases (e.g., DTLZ2 with three objectives), the algorithm misclassified a regular front as irregular, slightly degrading performance, an issue attributed to early-generation distribution noise.

Table 3 shows the HV results consistent with the IGD findings. NSGA-III-UR achieved the highest HV in 13 of 27 instances, with statistically significant wins in 10 comparisons against NSGA-III and 11 against A-NSGA-III. These outcomes confirm that NSGA-III-UR maintains a strong balance between convergence and diversity.

\begin{table}[ht]
\scalefont{0.82}
\renewcommand{\arraystretch}{1.2}
\centering
\caption{Hypervolume mean values obtained by NSGA-III, A-NSGA-III, and NSGA-III-UR on DTLZ1-7 and IDTLZ1-2 with 3 to 5 objectives.}
\begin{tabular}{@{}ccccc@{}}
\toprule
Problem&$M$&NSGA-III&A-NSGA-III&NSGA-III-UR\\
\midrule
\multirow{3}{*}{DTLZ1}&3&\hl{8.4585e-1 (8.67e-4) $\approx$}&8.3736e-1 (9.11e-3) $-$&8.4560e-1 (2.19e-3)\\
&4&9.3952e-1 (7.75e-4) $\approx$&9.3361e-1 (6.33e-3) $-$&\hl{9.3961e-1 (7.45e-4)}\\
&5&9.7113e-1 (5.33e-4) $\approx$&9.5926e-1 (9.94e-3) $-$&\hl{9.7119e-1 (4.93e-4)}\\
\hline
\multirow{3}{*}{DTLZ2}&3&\hl{5.6594e-1 (1.72e-5) $+$}&5.6390e-1 (1.96e-3) $-$&5.6497e-1 (1.41e-3)\\
&4&\hl{7.0514e-1 (5.35e-4) $\approx$}&7.0128e-1 (3.11e-3) $-$&7.0451e-1 (2.38e-3)\\
&5&\hl{7.7978e-1 (6.07e-4) $\approx$}&7.4875e-1 (3.20e-2) $-$&7.7973e-1 (3.95e-4)\\
\hline
\multirow{3}{*}{DTLZ3}&3&5.5080e-1 (5.94e-3) $\approx$&5.3934e-1 (1.33e-2) $-$&\hl{5.5237e-1 (9.00e-3)}\\
&4&\hl{6.7507e-1 (2.38e-2) $\approx$}&6.5842e-1 (2.04e-2) $\approx$&5.9860e-1 (2.00e-1)\\
&5&5.0388e-1 (3.17e-1) $\approx$&4.7420e-1 (3.20e-1) $-$&\hl{6.9173e-1 (1.22e-1)}\\
\hline
\multirow{3}{*}{DTLZ4}&3&5.3602e-1 (7.73e-2) $-$&5.2755e-1 (8.46e-2) $-$&\hl{5.4292e-1 (6.81e-2)}\\
&4&\hl{6.8914e-1 (4.77e-2) $\approx$}&6.7159e-1 (6.55e-2) $-$&6.8863e-1 (4.91e-2)\\
&5&7.4959e-1 (5.61e-2) $\approx$&\hl{7.6756e-1 (3.18e-2) $+$}&7.6335e-1 (4.27e-2)\\
\hline
\multirow{3}{*}{DTLZ5}&3&1.9599e-1 (7.78e-4) $-$&1.9692e-1 (6.68e-4) $\approx$&\hl{1.9699e-1 (4.67e-4)}\\
&4&1.3943e-1 (2.30e-3) $\approx$&\hl{1.4004e-1 (1.91e-3) $\approx$}&1.3979e-1 (2.63e-3)\\
&5&1.0265e-1 (9.66e-3) $\approx$&1.0444e-1 (6.16e-3) $\approx$&\hl{1.0677e-1 (5.63e-3)}\\
\hline
\multirow{3}{*}{DTLZ6}&3&1.9317e-1 (1.38e-3) $-$&\hl{1.9580e-1 (8.43e-4) $\approx$}&1.9547e-1 (6.26e-4)\\
&4&1.3118e-1 (7.69e-3) $\approx$&\hl{1.3203e-1 (3.69e-3) $\approx$}&1.3180e-1 (6.41e-3)\\
&5&\hl{9.4915e-2 (6.59e-3) $\approx$}&9.3729e-2 (5.57e-3) $\approx$&9.4166e-2 (5.40e-3)\\
\hline
\multirow{3}{*}{DTLZ7}&3&2.7494e-1 (1.07e-3) $-$&2.7502e-1 (1.20e-3) $-$&\hl{2.7570e-1 (9.57e-4)}\\
&4&\hl{2.5310e-1 (5.28e-3) $\approx$}&2.5209e-1 (4.52e-3) $\approx$&2.5227e-1 (4.44e-3)\\
&5&2.3940e-1 (3.74e-3) $\approx$&\hl{2.3946e-1 (2.84e-3) $\approx$}&2.3862e-1 (2.91e-3)\\
\hline
\multirow{3}{*}{IDTLZ1}&3&2.1260e-1 (1.67e-3) $-$&2.2351e-1 (2.23e-3) $\approx$&\hl{2.2444e-1 (1.21e-3)}\\
&4&3.8153e-2 (1.63e-3) $-$&4.6550e-2 (1.66e-3) $\approx$&\hl{4.6922e-2 (1.78e-3)}\\
&5&5.2951e-3 (5.35e-4) $-$&6.8912e-3 (1.15e-3) $\approx$&\hl{6.8921e-3 (9.30e-4)}\\
\hline
\multirow{3}{*}{IDTLZ2}&3&5.2580e-1 (2.49e-3) $-$&\hl{5.2856e-1 (2.90e-3) $\approx$}&5.2782e-1 (2.78e-3)\\
&4&2.2866e-1 (5.84e-3) $-$&2.4730e-1 (1.34e-2) $\approx$&\hl{2.4807e-1 (9.08e-3)}\\
&5&6.2711e-2 (4.54e-3) $-$&7.4986e-2 (6.87e-3) $\approx$&\hl{7.7059e-2 (4.58e-3)}\\
\hline
\multicolumn{2}{c}{$+/-/\approx$}&1/10/16&1/11/15&\\
\bottomrule
\end{tabular}
\label{HV_results}
\end{table}

In real-world scenarios (Table 4), NSGA-III-UR consistently outperformed both baselines in all four configurations. For MOKP, a combinatorial problem, and WRP, a continuous optimization problem, the proposed method demonstrated robustness and scalability, validating its applicability to heterogeneous domains.

Figure 3 illustrates representative POFs generated for DTLZ1 and IDTLZ1 under three and four objectives. The visual comparison confirms that NSGA-III-UR effectively retains the regular reference vector structure for regular problems and activates adaptation only in irregular cases.

\begin{table}[ht]
\scalefont{0.82}
\renewcommand{\arraystretch}{1.2}
\centering
\caption{Hypervolume mean values obtained by NSGA-III, A-NSGA-III, and NSGA-III-UR on real-world problems.}
\begin{tabular}{@{}ccccc@{}}
\toprule
Problem&$M$&NSGA-III&A-NSGA-III&NSGA-III-UR\\
\midrule
\multirow{3}{*}{MOKP}&3&3.2997e-1 (3.20e-3) $-$&3.2977e-1 (3.36e-3) $-$&\hl{3.3667e-1 (2.35e-3)}\\
&4&2.0429e-1 (2.37e-3) $\approx$&2.0411e-1 (2.54e-3) $\approx$&\hl{2.0494e-1 (2.61e-3)}\\
&5&1.1392e-1 (1.76e-3) $-$&1.1468e-1 (1.83e-3) $-$&\hl{1.1936e-1 (1.61e-3)}\\
\hline
\multirow{1}{*}{WRP}&5&9.6131e-2 (4.48e-4) $-$&9.6320e-2 (6.43e-4) $\approx$&\hl{9.6415e-2 (4.99e-4)}\\
\hline
\multicolumn{2}{c}{$+/-/\approx$}&0/3/1&0/2/2&\\
\bottomrule
\end{tabular}
\label{realworld_results}
\end{table}

\section{Conclusion}

In this paper, we have proposed NSGA-III with Update when Required (NSGA-III-UR) for MOPs and MaOPs. This model is based on the NSGA-III and A-NSGA-III frameworks combined with the update when required (UR) method. The UR method is used to verify the distribution of the population to decide whether given reference vectors must been adapted in the rest of the evolutionary process. Despite the use of the same framework as NSGA-III, the addition of checking the POF distribution to decide whether to use reference vector adaptation enables a competitive performance to be achieved when compared to its original peers in most MaOPs and MOPs with different PF shapes.

Despite the achievements, we can observe limitations in NSGA-III-UR, which raise possibilities for future lines of research:

\begin{enumerate}
    \item Using knowledge of the problem is interesting to select the appropriate parametric set. A study on POF classification methodologies can be investigated to increase the success rate in regular POFs. Furthermore, the behavior of the distribution of POFs in large-scale problems can be investigated; 

    \item Further investigation of the proposed algorithm to solve specific problem niches. The behavior and performance of the algorithm can be analyzed in constrained, high-dimensional, and dynamic optimization problems.
\end{enumerate}

\begin{figure}[ht]
\centering

\subfigure[DTLZ1 – 3 obj.]{\includegraphics[width=0.22\linewidth]{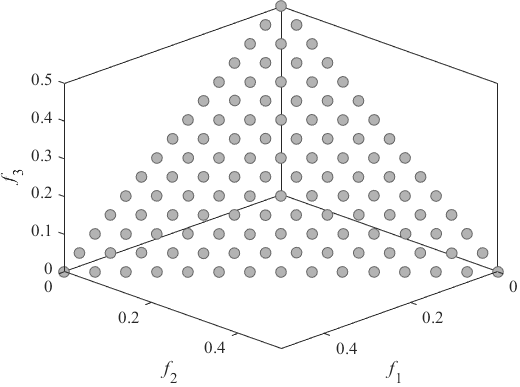}}
\subfigure[DTLZ1 – 4 obj.]{\includegraphics[width=0.22\linewidth]{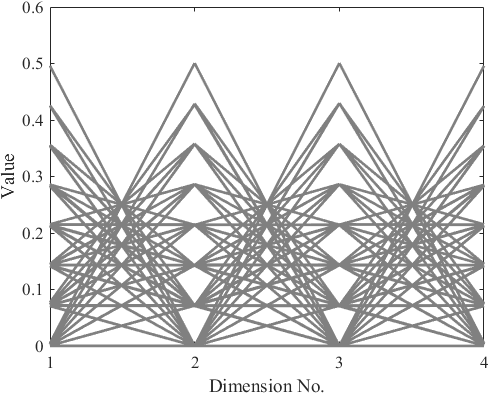}}
\subfigure[IDTLZ1 – 3 obj.]{\includegraphics[width=0.22\linewidth]{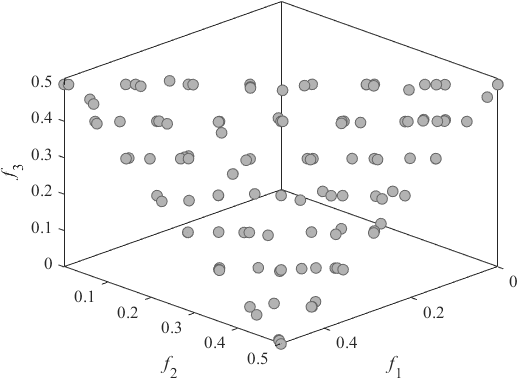}}
\subfigure[IDTLZ1 – 4 obj.]{\includegraphics[width=0.22\linewidth]{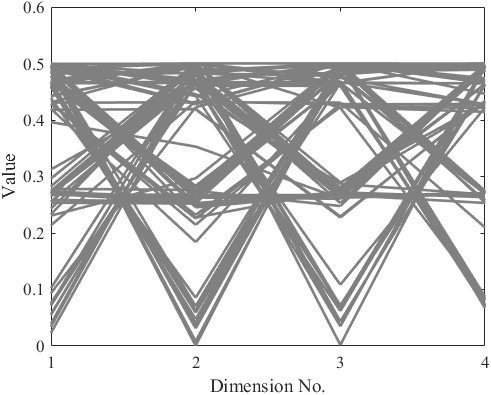}}

\subfigure[DTLZ1 – 3 obj.]{\includegraphics[width=0.22\linewidth]{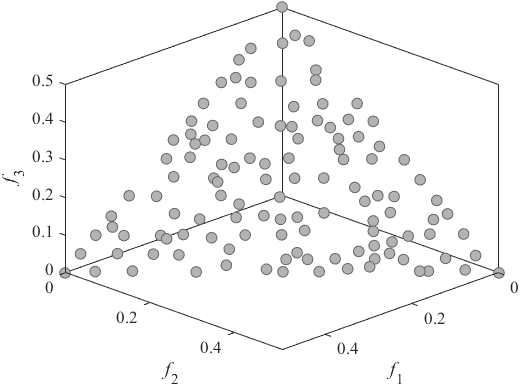}}
\subfigure[DTLZ1 – 4 obj.]{\includegraphics[width=0.22\linewidth]{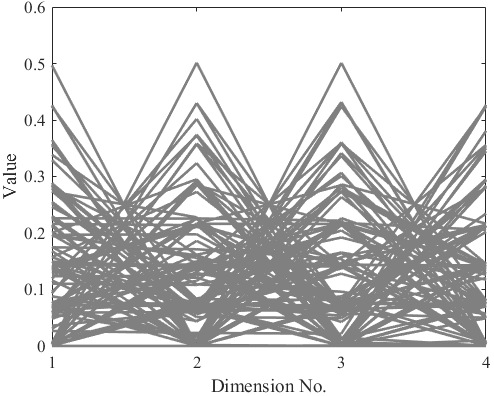}}
\subfigure[IDTLZ1 – 3 obj.]{\includegraphics[width=0.22\linewidth]{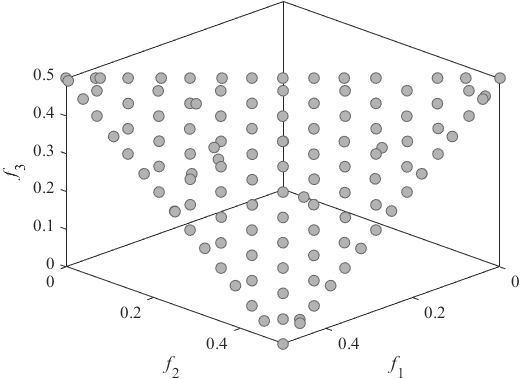}}
\subfigure[IDTLZ1 – 4 obj.]{\includegraphics[width=0.22\linewidth]{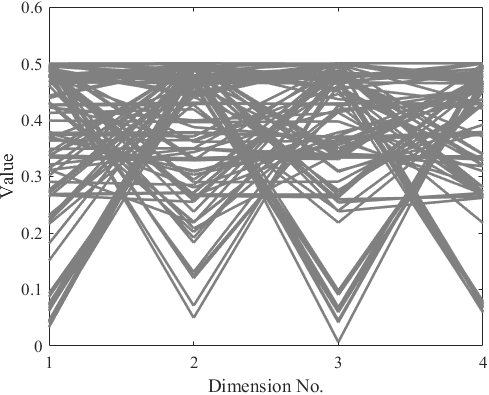}}

\subfigure[DTLZ1 – 3 obj.]{\includegraphics[width=0.22\linewidth]{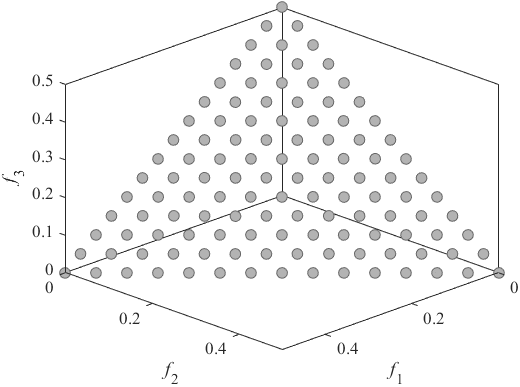}}
\subfigure[DTLZ1 – 4 obj.]{\includegraphics[width=0.22\linewidth]{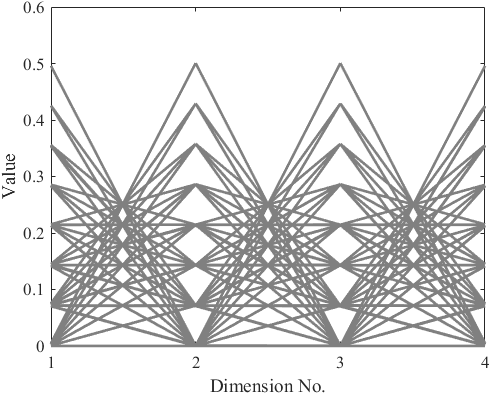}}
\subfigure[IDTLZ1 – 3 obj.]{\includegraphics[width=0.22\linewidth]{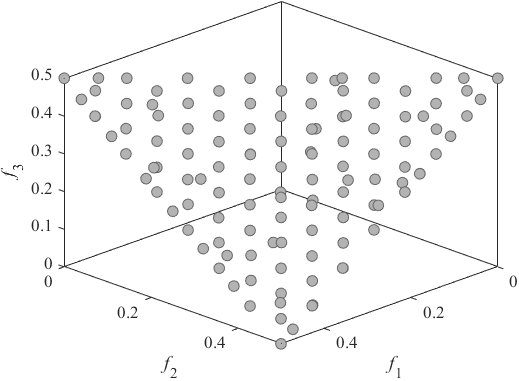}}
\subfigure[IDTLZ1 – 4 obj.]{\includegraphics[width=0.22\linewidth]{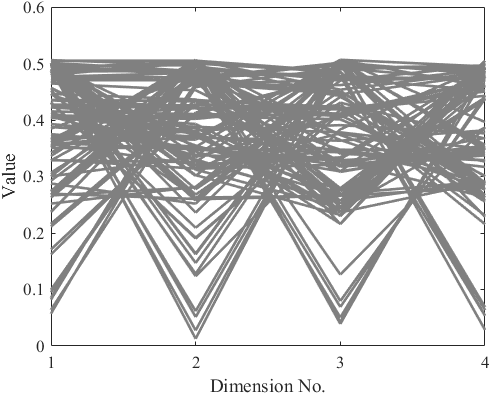}}

\caption{Final non-dominated solutions using the median IGD metric for DTLZ1 and IDTLZ1 with 3 and 4 objectives. Rows correspond to algorithms (top: NSGA-III, middle: A-NSGA-III, bottom: NSGA-III-UR); columns correspond to problem configurations.}
\label{fig:medianIGD}
\end{figure}

\bibliographystyle{sbc}
\bibliography{sbc-template}

\begin{thebibliography}{}

\bibitem[Awad et~al. 2022]{awad2022portfolio}
Awad, M., Abouhawwash, M., and Agiza, H.~N. (2022).
\newblock On nsga-ii and nsga-iii in portfolio management.
\newblock {\em Intelligent Automation \& Soft Computing}, 32(3):1893--1907.

\bibitem[Bosman and Thierens 2003]{bosman2003balance}
Bosman, P.~A. and Thierens, D. (2003).
\newblock The balance between proximity and diversity in multiobjective evolutionary algorithms.
\newblock {\em IEEE transactions on evolutionary computation}, 7(2):174--188.

\bibitem[Das and Dennis 1998]{das1998normal}
Das, I. and Dennis, J.~E. (1998).
\newblock Normal-boundary intersection: A new method for generating the pareto surface in nonlinear multicriteria optimization problems.
\newblock {\em SIAM Journal on Optimization}, 8(3):631--657.

\bibitem[de~Farias and Ara{\'u}jo 2022]{de2022decomposition}
de~Farias, L. R.~C. and Ara{\'u}jo, A. F.~R. (2022).
\newblock A decomposition-based many-objective evolutionary algorithm updating weights when required.
\newblock {\em Swarm and Evolutionary Computation}, 68:100980.

\bibitem[Deb and Jain 2014]{deb2014evolutionary}
Deb, K. and Jain, H. (2014).
\newblock An evolutionary many-objective optimization algorithm using reference-point-based nondominated sorting approach, part {I}: solving problems with box constraints.
\newblock {\em IEEE Transactions on Evolutionary Computation}, 18(4):577--601.

\bibitem[Deb et~al. 2005]{deb2005scalable}
Deb, K., Thiele, L., Laumanns, M., and Zitzler, E. (2005).
\newblock Scalable test problems for evolutionary multiobjective optimization.
\newblock In {\em Evolutionary Multiobjective Optimization}, pages 105--145. Springer.

\bibitem[Ishibuchi et~al. 2016]{ishibuchi2016performance}
Ishibuchi, H., Setoguchi, Y., Masuda, H., and Nojima, Y. (2016).
\newblock Performance comparison of {NSGA-II} and {NSGA-III} on various many-objective test problems.
\newblock In {\em IEEE Congress on Evolutionary Computation (CEC)}, pages 3045--3052.

\bibitem[Ishibuchi et~al. 2017]{ishibuchi2017performance}
Ishibuchi, H., Setoguchi, Y., Masuda, H., and Nojima, Y. (2017).
\newblock Performance of decomposition-based many-objective algorithms strongly depends on pareto front shapes.
\newblock {\em IEEE Transactions on Evolutionary Computation}, 21(2):169--190.

\bibitem[Jain and Deb 2013]{jain2013evolutionary}
Jain, H. and Deb, K. (2013).
\newblock An evolutionary many-objective optimization algorithm using reference-point based nondominated sorting approach, part {II}: Handling constraints and extending to an adaptive approach.
\newblock {\em IEEE Transactions on Evolutionary Computation}, 18(4):602--622.

\bibitem[Junqueira et~al. 2022]{junqueira2022moead}
Junqueira, P.~P., Meneghini, I.~R., and Guimar{\~a}es, F.~G. (2022).
\newblock Multi-objective evolutionary algorithm based on decomposition with an external archive and local-neighborhood based adaptation of weights.
\newblock {\em Swarm and Evolutionary Computation}, 71:101079.

\bibitem[Nuh et~al. 2021]{nuh2021performance}
Nuh, J.~A. et~al. (2021).
\newblock Performance evaluation metrics for multi-objective evolutionary algorithms in search-based software engineering: Systematic literature review.
\newblock {\em Applied Sciences}, 11(7):3117.

\bibitem[Sun et~al. 2024]{sun2024maoead}
Sun, Y., Liu, J., and Liu, Z. (2024).
\newblock Maoea/d with adaptive external population guided weight vector adjustment.
\newblock {\em Expert Systems with Applications}, 242:122720.

\bibitem[Tian et~al. 2017]{tian2017platemo}
Tian, Y., Cheng, R., Zhang, X., and Jin, Y. (2017).
\newblock Platemo: A matlab platform for evolutionary multi-objective optimization [educational forum].
\newblock {\em IEEE Computational Intelligence Magazine}, 12(4):73--87.

\bibitem[Zitzler and Thiele 1999]{zitzler1999multiobjective}
Zitzler, E. and Thiele, L. (1999).
\newblock Multiobjective evolutionary algorithms: a comparative case study and the strength pareto approach.
\newblock {\em IEEE Transactions on Evolutionary Computation}, 3(4):257--271.

\end{thebibliography}

\end{document}